\documentclass{article}
\pdfoutput=1
\usepackage[accepted]{arxivicml2015}
\usepackage{times}
\usepackage{graphicx}
\usepackage{subfigure}
\usepackage{natbib}
\usepackage{amssymb}
\usepackage{amsmath}
\usepackage{hyperref}
\usepackage[table]{xcolor}
\usepackage{rotating}

  \definecolor{mydarkblue}{rgb}{0,0.08,0.45}
  \hypersetup{ %
    pdftitle={},
    pdfauthor={},
    pdfsubject={Toxicity Prediction using Deep Neural Networks},
    pdfkeywords={},
    pdfborder=0 0 0,
    pdfpagemode=UseNone,
    colorlinks=true,
    linkcolor=mydarkblue,
    citecolor=mydarkblue,
    filecolor=mydarkblue,
    urlcolor=mydarkblue,
    pdfview=FitH}

\icmltitlerunning{Toxicity Prediction using Deep Learning}

\begin{document}

\twocolumn[
\icmltitle{Toxicity Prediction using Deep Learning}

\icmlauthor{Thomas Unterthiner$^*$ \footnotemark[1]$^{,}$\footnotemark[2]}{unterthiner@bioinf.jku.at}
\icmlauthor{Andreas Mayr$^*$ \footnotemark[1]$^{,}$\footnotemark[2]}{mayr@bioinf.jku.at}
\icmlauthor{G{\"u}nter Klambauer$^*$ \footnotemark[1]}{klambauer@bioinf.jku.at}
\icmlauthor{Sepp Hochreiter \footnotemark[1]}{hochreit@bioinf.jku.at}
\icmladdress{\footnotemark[1] Institute of Bioinformatics, Johannes Kepler University Linz, Austria}
\icmladdress{\footnotemark[2] RISC Software GmbH, Johannes Kepler University Linz, Austria}
\icmladdress{$^*$These authors contributed equally to this work}

\icmlkeywords{toxicity prediction, tox21, virtual screening, deep learning, multitask learning, machine learning, ICML}

\vskip 0.3in
]

\begin{abstract}
Everyday we are exposed to various chemicals
via food additives, cleaning and cosmetic products and
medicines --- and some of them might be toxic.
However testing the toxicity of all existing compounds
by biological experiments is neither financially nor logistically feasible.
Therefore the government agencies NIH, EPA and FDA launched
the Tox21 Data Challenge within
the ``Toxicology in the 21st Century'' (Tox21) initiative.
The goal of this challenge was to assess the
performance of computational methods in predicting the toxicity
of chemical compounds.
State of the art toxicity prediction methods build upon specifically-designed
chemical descriptors developed over decades.
Though Deep Learning is new to the field and was never applied to toxicity prediction before,
it clearly outperformed all other participating methods.
In this application paper we show that
deep nets automatically learn features
resembling well-established toxicophores.
In total, our Deep Learning approach won
both of the panel-challenges (nuclear receptors and stress response)
as well as the overall Grand Challenge, and thereby sets a new standard in tox prediction.
\end{abstract}

\section{Introduction}

Throughout their lives people are exposed to a sheer endless variety of chemical compounds,
many of which are potentially dangerous.
Determining the toxicity of a chemical is of crucial importance in order to minimize our exposure to
harmful substances in every day products.
Toxicity is also a central issue in the development of new drugs,
with more than 30\,\% of drug candidates failing in clinical trials
because of undetected toxic effects\,\cite{bib:Kola2004, bib:Arrowsmith2011}.

In 2008, the U.\,S. National Institutes of Health (NIH) and the U.\,S.
Environmental Protection Agency (EPA), agreed on collaborating on future
toxicity testing activities\,\cite{bib:tox212007}. Their efforts were later
joined by the U.\,S. Food and Drug Administration (FDA) under the umbrella of
the \emph{Tox21 Program}. The program's stated goals are to develop better toxicity
assessment methods, as current methods are not likely to scale with the
increased demand for effective toxicity testing.

Current methods for testing the toxicity of a high number of chemicals rely on
High-Throughput Screening (HTS).
HTS experiments can investigate whether a chemical compound at a given concentration
exhibits a certain type of toxicity, for a number of different compounds in parallel.
These experiments are repeated with varying concentrations of the chemical
compound, which allows to determine dose-response curves\,\cite{bib:Inglese2006}.
From these curves one can reliably determine whether a compound activated a
given pathway or receptor, inhibited it or did not interact at all.

Conducting these HTS experiments is a time- and cost-intensive process.
Typically, a compound has to be tested for several types of toxicity at
different concentration levels. Thus, the whole procedure has to be rerun for
many times for each compound. Usually, a cell line has to be cultivated
to obtain a single data point.
Even an unprecedented multi-million-dollar effort, the Tox21 project, could test
only a few thousands of compounds for as few as twelve toxic effects.
Therefore, accurate computational methods for accurate prediction of toxic effects are
highly demanded.



Existing computational approaches can be grouped into structure- and ligand-based.
The structure-based methods simulate physical interactions between the compound
and a biomolecular target\,\citep{bib:Kitchen2004} but are only applicable
if the complete 3D structure of all interacting molecules are known, and they are
infeasible for larger compound data bases.  
Ligand-based approaches predict the interactions based on previous
measurements\,\citep{bib:Jenkins2007}. Previous machine learning efforts
were almost always ligand-based, such as scoring
approaches like the Naive Bayes statistics\,\citep{bib:Xia2004, bib:Nigsch2008, bib:Mussa2013},
density estimation\,\citep{bib:Lowe2012, bib:Harper2001},
nearest neighbor, support vector machines, and shallow feed forward neural networks
\,\citep{bib:Byvatov2003, bib:Lowe2011}.

In 2012, the Merck Kaggle challenge on chemical compound activity was won
using deep neural networks, and the winning group later showed that multi-task
learning can help to predict biological activities on single proteins\,\cite{bib:Dahl2014}.
Dahl's success inspired us to use Deep Learning for toxicity and target prediction\,\cite{bib:unterthiner2014}.
In contrast to biological activities of proteins, toxicological
effects involve whole cell states determined by
dysregulated biological processes. More specifically, toxicity prediction mainly focuses on cellular assays
which measure cytotoxicity, i.e., they measure if a compound is toxic
to a cell. A (cyto)toxic compound will cause harm to a cell, e.g. by causing
acute mechanical injury or by triggering the programmed cell death mechanism (apoptosis)
in the affected cells, which multicellular organisms use to protect themselves from cells
that have gone out of control.

\subsection{Deep Learning for Toxicity Prediction}
Deep learning architectures seem to be well suited for toxicity prediction
because they (1) automatically
construct complex features \cite{bib:Bengio2013a} and (2) allow for multi-task
learning \cite{bib:Caruana1997,bib:Deng2013,bib:Bengio2013a}.

One key aspect of toxicological research is its reliance on hierarchical levels
of abstraction when thinking about chemical structures. A major research goal
is the identification of toxicophores,\,\cite{bib:Kier1971, bib:Lin2000} 
which are the sets of steric and electronic properties that together produce
a certain toxicological effect. These properties include hydrophobic regions,
aromatic rings, electron acceptors or donors.

\begin{figure*}
    \begin{center}
    \includegraphics[width=0.7\textwidth]{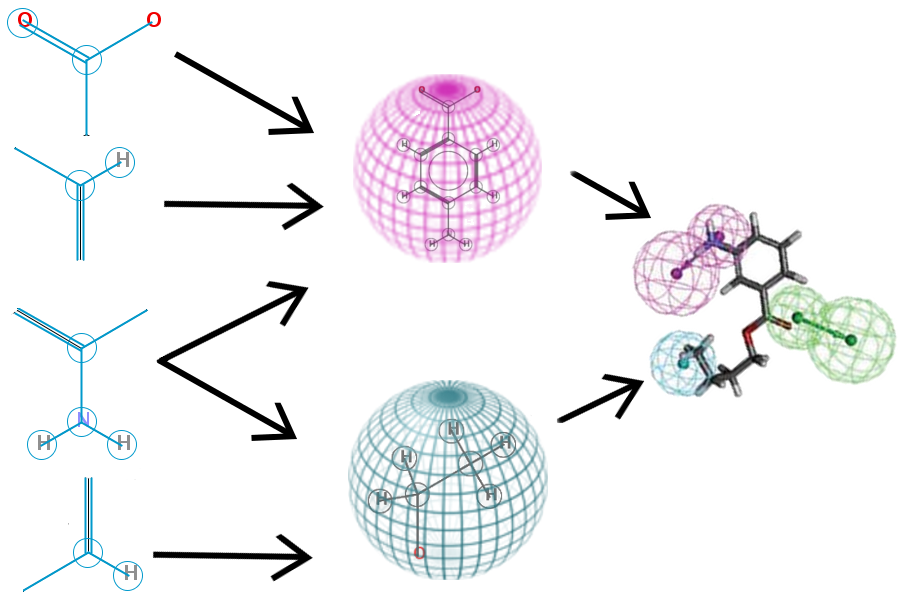}
    \end{center}
    \caption{Hierarchical nature of fingerprint features: by combining
    the ECFP features we can build reactive centers. By pooling specific
    reactive centers together we obtain a toxicophore that encodes
    a specific toxicological effect.}
    \label{fig:toxicophore}
\end{figure*}

This maps naturally to Deep Learning architectures, where higher levels represent
more complex concepts \cite{bib:Bengio2013b}. This idea is
depicted in \autoref{fig:toxicophore}, where  ECFP4 input data
(chemical substructures) represent low level properties in their first layer,
which are combined to form reactive centers, which in turn encode toxicophores in higher layers.

Additionally, Deep Learning is ideally suited for multi-task learning,
which is a common setting for toxicology prediction:
The same compound is often under investigation for several types of toxicity, and each of
these types is its own prediction task. The work of \cite{bib:Ramsundar2015} also
shows that the multi-task environment does help when predicing chemical compounds,
and that the performance boost obtained this way increases with the number
of additional learning tasks. However, we typically
have to deal with missing labels, as not all compounds will have been tested
for each type of toxicity, or because some measurements were inconclusive.

Integrating all prediction tasks into one overarching multi-task setting offers two advantages:
(a) it naturally allows for multi-label information and therefore
can utilize relations between tasks;
(b) it allows to share hidden unit representations among prediction tasks.
The latter item is particularly important in our application as for some tasks very
few measurements are available, therefore single-task prediction may fail
to construct an effective representation.
Thus, deep networks exploit representations learned across different tasks and
can boost the performance on tasks with few training examples.
Furthermore, this method allows us to predict an arbitrary number of toxicological
effects at the same time, without the need to train single classifiers for each one.



\section{Methods} \label{lbl:methods}

\subsection{DNN Architecture}

Our system takes a numerical descriptor of a given compound as input, and tries
to predict several different types of toxic effects
at the same time. Such a type
could be e.g. whether the compound acts as inhibitor to a specific
nuclear receptor, or whether it activates a specific stress response pathway. Each
of these types is a binary prediction task.

Formally, the problem we are trying to solve presents itself as follows: given
a chemical compound $i$, we want to predict whether the compound has property $t$.
We encode this information in the binary value $y_{it}$, where $y_{it} = 1$ if the
compound has the property and $y_{it} = 0$ otherwise.
We are interested in predicting the behavior of a compound on $T$ properties at
the same time.

Each compound is represented using a number of numerical (or binary) features described
later in this section. As training data, we are given a numerical
representation $\mathbf{x_i} \in \mathbb{R}^d$ of $n$ training
compounds as well as a sparsely populated
matrix $\mathbf{Y} \in \mathbb{R}^{n \cdot m}$ of  measurements.

We solve this by using a training objective
that is the weighted sum of the cross-entropies over all tasks $t$:
\begin{align*}
- \sum_t^T m_{ti} \left( y_{it} \log\left( \sigma_t({\bf x_i})\right) + (1 - y_{it}) \log\left(1 - \sigma_t({\bf x_i})\right)\right)
\end{align*}

The binary variable $m_{ti}$ is 1 if sample $i$ has a valid label for task $t$
and 0 otherwise. Each single training sample contributed only to a few of
the tasks. Thus, output units that were not active during a training sample
were masked during backpropagation by multiplying their $\delta$ error by $m_{ti}$.

Our network consists of one or multiple layers of ReLU hidden
units\,\citep{bib:Nair2010, bib:Glorot2011},
followed by one layer of one or more sigmoid output units, one for
each classification task.

\subsection{Hyperparameters}\label{lbl:hyperparameters}
The input features had substantially different scales and distributions, such that
it was not obvious how to best preprocess them. We tried
both the standard deviation as well as simple $\tanh$ nonlinearity to bring
the chemical descriptors in the same range. ECFP4 features were either scaled
by $tanh$ or sqrt nonlinearities. We additionally used a simple thresholding
scheme to filter out very sparse features, which helped to bring the number of
features down into a manageable range.

We tried different combinations of the available features, e.g. using
only the binary ECFP4 fingerprints, or combining only the chemical descriptors
with the toxicophore features.

To regularize our network, we used both Dropout\,\cite{bib:Hinton2012, bib:Srivastava2014}
as well as small amounts of L2 weight decay, which both work in concert to avoid regularization
\,\cite{bib:Krizhevsky2012, bib:Dahl2014}. Additionally, we used Early Stopping
as determined via cross-validation.

\autoref{table:hyperparams} contains the complete list of hyperparameters we
used for our network, as well as the search range for each parameter.

\begin{table*}[!ht]
\begin{center}
\begin{tabular}{ll}
\multicolumn{1}{c}{\bf Hyperparameter}  &\multicolumn{1}{c}{\bf Considered values}\\
\\
  Normalization & \{standard-deviation, tanh, sqrt\} \\
  Feature type & \{molecular-descriptors, tox-and-scaffold-similarities, ECFP4\} \\
  Fingerprint sparseness threshold & \{5, 10, 20\} \\
  Number of Hidden Units & \{1024, 4096, 8192, 16356\} \\
  Number of Layers & \{1, 2, 3\} \\
  Learning Rate & \{0.01, 0.05, 0.1\} \\
  Dropout & \{no, yes (50\% Hidden Dropout, 20\% Input Dropout)\} \\
  L2 Weight Decay & \{0, $10^{-6}$, $10^{-5}$, $10^{-4}$\}
\end{tabular}
\end{center}
  \caption{Hyperparameters considered for the neural networks.
  {\bf Normalization:} Scaling of the predefined features.
  {\bf Feature type:} Determines which of the
  features were used as input features. ``molecular-descriptors'' were the
  real-valued descriptors. ``tox-and-scaffold-similarities'' were the similarity
  scores to known toxicophores and scaffolds, ``ECFP4'' were the ECFP4 fingerprint
  features. We tested all possible combinations of these features.
  {\bf Fingerprint sparseness threshold:} A feature was not used if it was only
  present in fewer compounds than the given number.
  {\bf Number of hidden units:} The number of units in the hidden layer of
  the neural network.
  {\bf Number of layers:} The number of layers of the neural network.
  {\bf Learning rate:} The learning rate for the backpropagation algorithm.
  {\bf Dropout:} Dropout rates.
  {\bf L2 Weight Decay:} The weight decay hyperparameter.
  }
  \label{table:hyperparams}
\end{table*}

\subsection{Input Features}\label{lbl:inputfeatures}
Having good input features is a crucial issue for chemoinformatics applications.
A vast variety of different methods exist, which calculate numerical features
of the the typical graph-based storage format used for chemical compounds.

We used a high-dimensional binary representation using
Extended Connectivity FingerPrint (ECFP4) features, the currently
best performing compound description in drug design applications\,\cite{bib:Rogers2010}.
Each feature/fingerprint denotes the presence-count
 of a certain chemical substructure, such as the ones given
on the left-most column of \autoref{fig:toxicophore}.
In total, this produced approximately 30\,000 very sparse features.
As part of the hyperparameter selection we used a sparsity filter to
emove non-informative ones.

We also calculated the similarity of each compound to 2\,500 known toxicophore
features, ie., patterns of substructures that were previously reported as
toxicophores in the literature\,\cite{bib:Kazius2005}.
We also calculated the similarity of each compound with 200 common
chemical substructures that appear often in organic molecules.

Additionally, we calculated a number of descriptors based on the topological
and physical properties of each compound. Typical descriptors for toxicity
prediction can be grouped into 1D, 2D and 3D features\,\cite{bib:Hong2008}.
Features that revolve around scalar properties such as counts of occurences
for various atom-types, molecular weight or size are 1D features, while
2D features can be extracted from the planar chemical structure graph. These
include graph-based features, 2D autocorrelation descriptors
 as well as van der Waals volume or the sum of Pauling
atomic polarizabilities. Finally 3D structures usually involve force-field
and quantum-mechanical simluations to extract things like solvent accessible surface area
or partial charge informations.

We calculated a variety of these descriptors using off-the-shelf software\,\cite{bib:Cao2013b}.
However, not all descriptors could be calculated for all compounds.
We used median-imputation to deal with missing values whenever feasible.
This way we obtained a total of 5057 additional features.

\subsection{Implementation}
Depending on hyperparameter settings, our deep neural network had to deal with
up to 40\,000 input features and very large hidden layers. We stored the weight
parameters on a single GPU with 12\,GB RAM and used mini-batches of
512 samples 
for stochastic gradient descent learning.
Since storing our input data in dense format requires about 5\,TB of
disk space, we used a sparse storage format. However, it proved to be faster to
upload a mini-batch in sparse format to the GPU and then convert it to dense format
instead of using sparse matrix multiplication.

\section{Experimental Results}\label{lbl:experiments}
\subsection{Tox21 Data Challenge Data}
We validated our approach using the data from the Tox21 Data
Challenge\,\cite{bib:tox21challenge}, a toxicity prediction challenge
organized by the Tox21 program partners open to participants worldwide. The
data for this challenge was collected within the framework of the Tox21 research
initiative, which aims to produce highly realiable measurements with
stringent quality-control criteria, that are otherwise hard to come by in public databases.

The data set provided by the Tox21 Data Challenge included approximately 12\,000
compounds and was composed of twelve different sub-challenges/tasks. Each sub-challenge
required the prediction of a different type of toxicity.
The sub-challenges were split between two panels:
Seven of the twelve sub-challenges dealt with \emph{Nuclear Receptor} (NR) signaling pathways,
the remaining five with the \emph{Stress Response} (SR) pathways.

Nuclear receptors are important components in cell communication and control,
and are involved in development, metabolism and proliferation. They have been
shown to play a key role in toxicology as well\,\cite{bib:Woods2007}.
The Tox21  data set investigated several NRs involved in endocrine system,
i.e., the secretion of hormones into the blood stream, as toxins can cause
disruption of the normal endocrine function. Two such nuclear hormone receptors,
the estrogen and the androgen receptor, have been measured by two independent
systems, once using a luminescence method, and once using a modified
antibiotic resistance gene (\emph{NR.ER} and \emph{NR.ER.LBD} / \emph{NR.AR} and
\emph{NR.AR.LBD} respectively). Furthermore,
the challenge included a task on predicting the antagonists of the aromatase
enzyme, which catalyzes the conversion of androgen to estrogen and thereby keeps
the balance between these two hormones (\emph{NR.Aromatase}).
The last two NRs in the Tox21 data set were the aryl hydrocarbon receptor
(\emph{NR.AhR}) which is essential for reacting to a cell's environmental changes,
and a specific subtype of the peroxisome proliferator-activated receptors (\emph{NR.PPAR.gamma})
which is involved in the regulation of various genes as well as metabolism. Overall
the NR tasks included a broad variety of different toxicity-related receptors.

\begin{figure}[ht]
    \centering
        \includegraphics[width=\columnwidth] {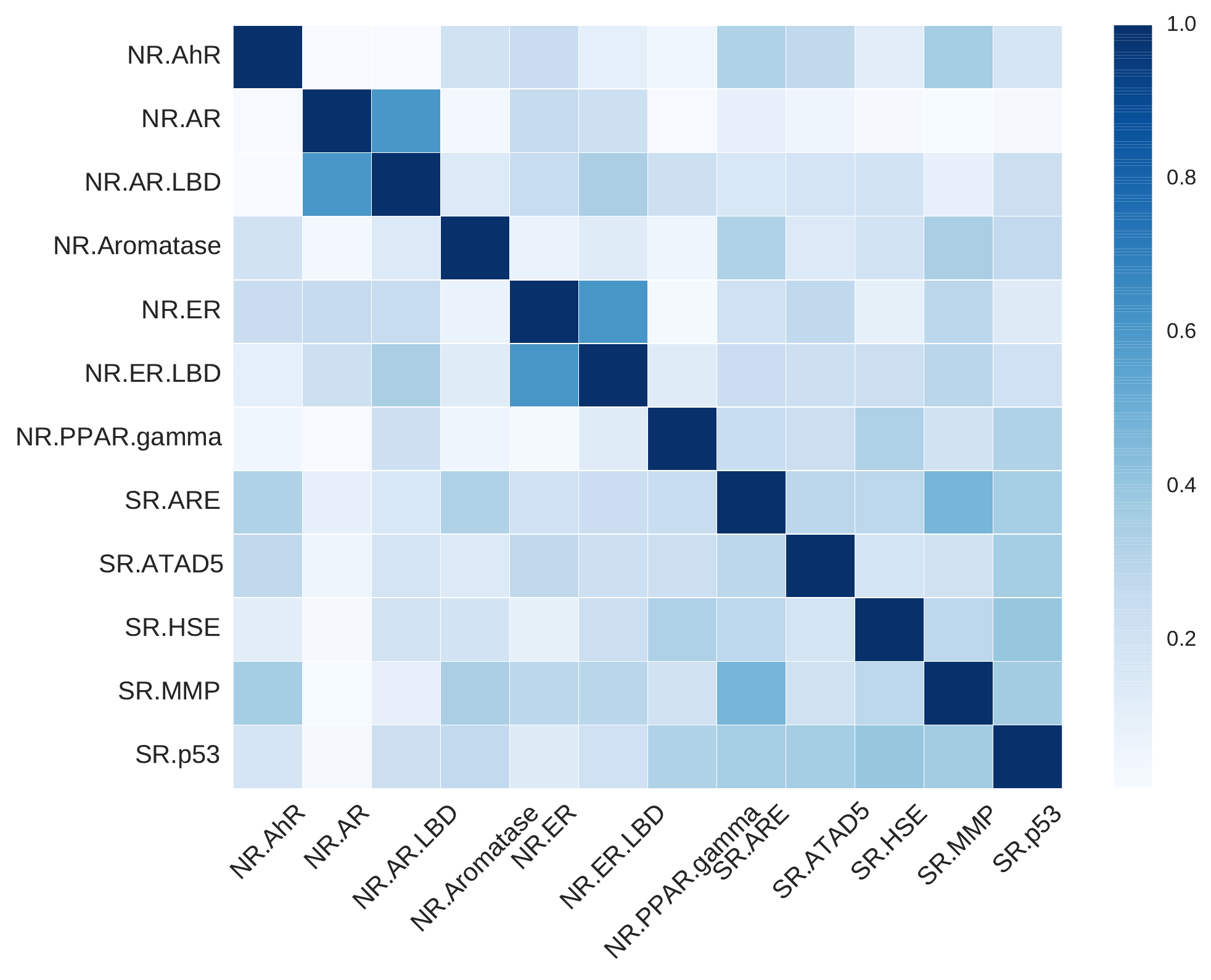}
\caption{
    Absolute correlation coefficient between the different tasks of the Tox21
    Data Challenge}
\label{fig:correlation}
\end{figure}

Toxicity can also cause cellular stress which in term can lead to apoptosis. Therefore
the Tox21 data also includes five tasks on various stress response indicators:
The antioxidant response element signaling pathway (\emph{SR.ARE}) directly reacts to oxidative stress,
while the heat shock factor response element (\emph{SR.HSE}) is involved in reacting
to heat shocks as part of the cell's internal repair mechanisms. The
ATAD5 signaling pathway will be activated when a cell detects DNA damage
(\emph{SR.ATAD5}). The SR panel also includes a task
on predicting which compounds influence the mitochondrial membrane potential
(\emph{SR.MMP}), which is essential for generating the energy a cell consumes.
Finally, the p53 task requires participants to detect activation of the p53 pathway (\emph{SR.p53}),
a well known cancer pathway which is activated both by DNA damage, but also reacts to
 various other cellular stresses.  For this reason, a compound that triggers
 any of the other stress response pathways has a high probability to also show
up as active on the p53 task. In general, all of the SR tasks show higher correlation
with each other than the nuclear receptor tasks (c.f. \autoref{fig:correlation}).


%
%
%
%

\begin{figure}[ht]
    \centering
        \includegraphics[width=\columnwidth] {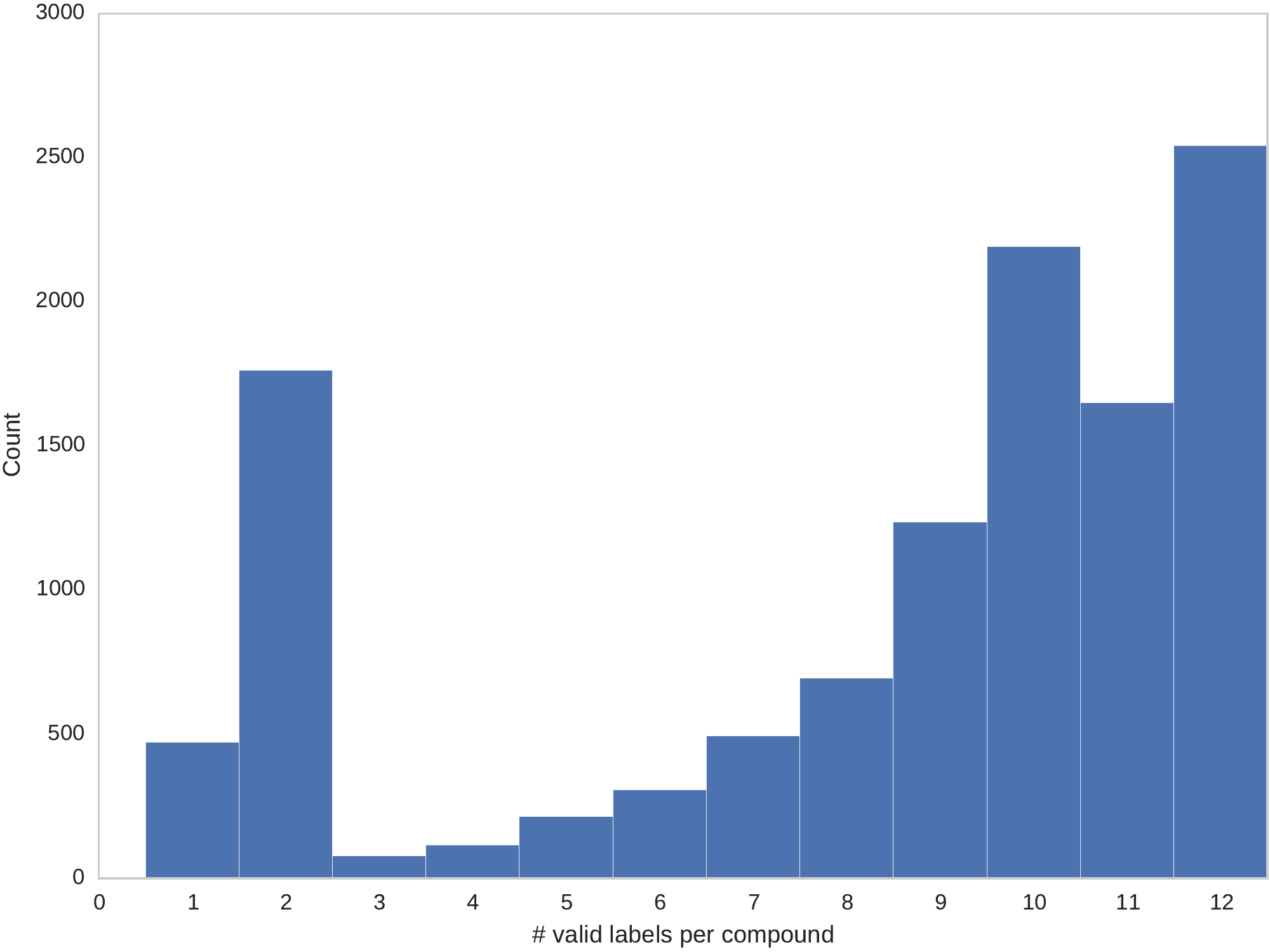}
\caption{
    Number of tasks each training compound of the Tox21 Data Challenge was
    part of. Only $\approx 500$ compounds were part of only a single task, with
    over half (54\,\%) of the compounds being labeled in 10 or more of the tasks.
}
\label{fig:labelspercompound}
\end{figure}

Most of the compounds were measured on several of the tasks (c.f. \autoref{fig:labelspercompound}),
such that all the tasks operated on subsets of the same overall data set. This
allowed us to compute correlations between the tasks, displayed in \autoref{fig:correlation}.
As expected, the tasks that involved measuring the same pathway via different
methods (AR/AR-LBD and ER/ER-LBD) were highly correlated.  Also, the p53
pathway, which is one of the main focal points of stress response signaling,
showed high levels of correlation with the other tasks that measured specific
stress responses.

Overall, the compounds were split into a training set consisting of 11\,764 compounds
with known labels, a leaderboard set used to rank participants on a public leaderboard (297
compounds) as well as a private test set used for the final evaluation of all
submitted entries (643 compounds).
The labels for the leaderboard set were initially held back, but later made available
to the participants in the final stages of the competition, while
the labels of the final test set have not yet been released.

\begin{table}[htb]
\centering
\begin{tabular}{l|ccc}
{\bf Task} & {\bf AUC ST} & {\bf AUC MT} & {\bf $p$-value}\\
\hline
       NR.AhR &  0.8487 & 0.8409	& 0.072 \\
        NR.AR &  0.3755 & 0.3459	& 0.202 \\
    NR.AR.LBD &  0.8799 & {\bf 0.9289}	& 0.011 \\
 NR.Aromatase &  0.7523 & {\bf 0.7921}	& 0.006 \\
        NR.ER &  0.6659 & {\bf 0.6949}	& 0.006 \\
    NR.ER.LBD &  0.6532 & {\bf 0.7272}	& 0.006 \\
NR.PPAR.gamma &  0.6367 & {\bf 0.7102}	& 0.006 \\
       SR.ARE &  0.7927 & 0.8017	& 0.148 \\
     SR.ATAD5 &  0.7972 & 0.7958	& 0.338 \\
       SR.HSE &  0.7354 & {\bf 0.8101}	& 0.006 \\
       SR.MMP &  0.8485 & 0.8489	& 0.265 \\
       SR.p53 &  0.6955 & {\bf 0.7487}	& 0.006 \\
\end{tabular}
\caption{Comparing single-task (ST) and multi-task (MT) learning.
Evaluation was done on the Tox21 leaderboard set.
Results are the mean values of training 5 nets from different random initializations.
Significant differences according to a two-sided Mann - Whitney U test in bold.}
\label{table:st-vs-mt}
\end{table}

\subsubsection{Data Preprocessing}
The Tox21 training set contains redundant compounds that appear multiple times within
the data, but each time accompanied by carrier molecules such as water, salts
or other solubles. Also, we observed compounds that actually consisted
of two unrelated structures, but which for some unknown reason where encoded together.
We semi-automatically labeled these fragments, cleaning up contradictory and
combining agreeing compounds. This way we identified 8,695 distinct compound
fragments.

To further clean up the data, we made ran a standard clean-up routine for chemical
compounds on the data using ChemAxon. 
This made all hydrogen atoms explicit, ensured that aromatic bonds and tautomers where coded
consistently and unified the encoding of salts.
We then calculated the input features as described in \autoref{lbl:inputfeatures}.

\begin{figure*}[h!t]
    \centering
        \includegraphics[width=0.3\textwidth] {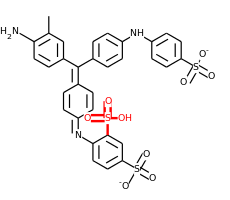}
        \includegraphics[width=0.32\textwidth] {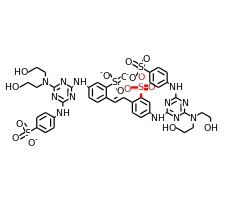}
        \includegraphics[width=0.3\textwidth] {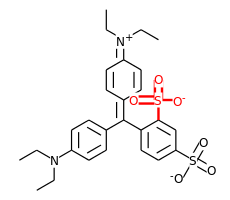} \\
        \rule[1ex]{\textwidth}{0.5pt}
        \includegraphics[width=0.3\textwidth] {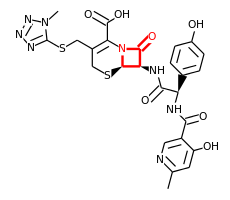}
        \includegraphics[width=0.3\textwidth] {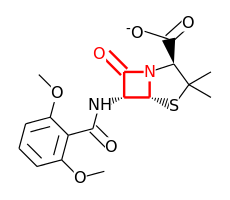}
        \includegraphics[width=0.3\textwidth] {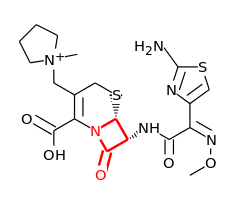} \\
        \rule[1ex]{\textwidth}{0.5pt}
        \includegraphics[width=0.33\textwidth] {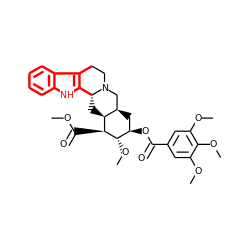}
        \includegraphics[width=0.33\textwidth] {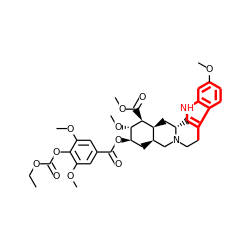}
        \includegraphics[width=0.33\textwidth] {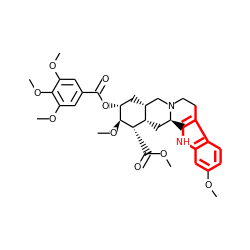}
\caption{Neurons that have learned to detect the presence of pharmacophores.
Each row shows a certain hidden unit in a learned network that correlates highly with a certain
toxicophore feature. The row shows the three chemical compounds that had the highest
activation for that neuron. Emphasized in red is the toxicophore structure
from the literature that the neuron correlates with. Rows 1 and 2 are from the
first hidden layer, the last row is from a higher layer.}
\label{fig:toxicophorestructures}
\end{figure*}

\subsection{Evaluation}
We defined cross-validation sets for hyperparameter selection, optimizing for two goals:
a) The class-distributions should be close to the final test set.
In the training set many compounds were only measured on a small subset of assay, whereas
we expected compounds in the final test set to be labeled on all twelve
tasks. We therefore included only compounds that were labeled on at least eight
tasks in the cross-validation sets. The remaining, sparsely labeled compounds
were added to the training set of each fold.
b) The cross-validation sets should not be overly simple. We wanted to avoid the
situation where the training samples were exceedingly similar to the test samples. This
happens frequently within chemical data because a number of compounds might share the same
chemical backbone. Therefore, we clustered the compounds according to their
structural similarity\,\cite{bib:Klambauer2015}
and distributed the resulting clusters among the five cross-validation folds.

We used the AUC score as quality criterion, which we optimized independently for each task.
So even though we employed multi-task networks, we optimized the hyperparameters
differently for each task at hand.

\subsection{Multitask Learning}

Most of the compounds where labeled on several of the tasks
(c.f. \autoref{fig:labelspercompound}), which allowed us to calculate the
correlation between different tasks. As can be
seen in \autoref{fig:correlation}, the twelve different task of the Tox21 Data Challenge Data
 were highly correlated with one another. Thus, this was an ideal setting for
multi-task learning.

To see whether multi-task learning really helps in this scenario as much as it
did when predicting biological activities on protein level\,\cite{bib:Dahl2014},
we also trained single-task neural networks on the same tasks.

As shown in \autoref{table:st-vs-mt}, in almost all tasks the multi-task learning
approach significantly outperforms the single task networks. Both networks failed
in one task which suffered from very unbalanced class distribution (only 3
positive examples in the leaderboard set).

\subsection{Learning Toxicophore Representation}
One of the hallmarks of Deep Learning are several layers of
hierarchical representations of increasing abstractions\,\cite{bib:Bengio2013a}.
Within the chemical research community such a hierarchy of features has naturally
emerged: single atoms are grouped together as functional groups and reactive
centers, which in turn define \emph{toxicophores} (c.f. \autoref{fig:toxicophore}.
Such features are the state-of-the-art way that chemists
and drug designers think about the properties of each chemical compound\,\cite{bib:Kazius2005}.
To determine the effectiveness of Deep Learning for toxicity prediction, we
investigated whether the network did implicitly encode toxicophore features in
its hidden layers.

We trained a multi-task deep network on the Tox21 data using exclusively ECFP4
fingerprint-features as input. Each fingerprint encodes how many times a
specific, small chemical substructure appears within a compound. No other
input features were used.

After training, we computed the correlation between the activations of the
hidden units and the presence/absence of known toxicophore features in the
compounds. We did indeed find several highly significant correlations, clearly
demonstrating that the hidden units of a neural network do indeed automatically
learn toxicophore structures.

Visual inspection of the results showed that lower layers did tend to learn
smaller features, often focusing on single functional groups like
e.g. sulfonyl-groups (see row 1 and 2 of \autoref{fig:toxicophorestructures}, while
in higher layers the correlations were more with larger toxicophore clusters,
even involving structures that did not match the toxicophore perfectly
(row 3 of \autoref{fig:toxicophorestructures}.

\section{Results}
The Tox21 Data Challenge Data attracted a large crowd of participants from
all over the world, including submissions from leading research labs and
industry.

The final evaluation was done by the organizers on a held back evaluation set
consisting of 643 compounds. The teams were allowed to send in predictions
for these final compounds, but did not receive any feedback as to how well
they fared. The final scoring on each sub-challenge was based on the AUC values
of each team's final submission.

\newcommand{\rankone}[1]{\cellcolor{gray!35}\textbf{#1}}
\newcommand{\ranktwo}[1]{\cellcolor{gray!20}#1}
\newcommand{\rankthree}[1]{#1}

\newcommand{\rot}[1]{\begin{rotate}{90}#1\end{rotate}}

\begin{table*}[!htbp]
\centering
\small
\begin{tabular}{l@{\hskip 0.1in}r@{\hskip 0.1in}r@{\hskip 0.1in}r@{\hskip 0.1in}||r@{\hskip 0.1in}r@{\hskip 0.1in}r@{\hskip 0.1in}r@{\hskip 0.1in}r@{\hskip 0.1in}r@{\hskip 0.1in}r@{\hskip 0.1in}r@{\hskip 0.1in}r@{\hskip 0.1in}r@{\hskip 0.1in}r@{\hskip 0.1in}r@{\hskip 0.1in}}
 \\
 \\
 \\
 \\
 & {\bf \rot{AVG}} & {\bf \rot{NR}} & {\bf \rot{SR}} & {\bf \rot{AhR}} & {\bf \rot{AR}} & {\bf \rot{AR-LBD}} & {\bf \rot{ARE}} & {\bf \rot{Aromatase}} & {\bf \rot{ATAD5}} & {\bf \rot{ER}} & {\bf \rot{ER-LBD}} & {\bf \rot{HSE}} & {\bf \rot{MMP}} & {\bf \rot{p53}} & {\bf \rot{PPAR.g}} \\
\textbf{\emph{our method}} & \rankone{0.846} & \rankone{0.826} & \rankone{0.858} & \rankone{0.928} & \ranktwo{0.807} & \rankone{0.850} & \rankone{0.840} & \ranktwo{0.834} & 0.793 & \ranktwo{0.793} & \ranktwo{0.814} & \ranktwo{0.858} & 0.941 & \ranktwo{0.862} & \rankone{0.839} \\
\textbf{AMAZIZ} & \ranktwo{0.838} & \ranktwo{0.816} & \ranktwo{0.854} & \ranktwo{0.913} & 0.770 & 0.846 & \ranktwo{0.805} & 0.819 & \rankone{0.828} & \rankone{0.806} & 0.806 & 0.842 & \rankone{0.950} & 0.843 & 0.830 \\
\textbf{dmlab} & 0.824 & 0.811 & 0.850 & 0.781 & \rankone{0.828} & 0.819 & 0.768 & \rankone{0.838} & 0.800 & 0.766 & 0.772 & \rankthree{0.855} & \ranktwo{0.946} & \rankone{0.880} & \ranktwo{0.831} \\
\textbf{T} & 0.823 & 0.798 & 0.842 & 0.913 & 0.676 & \ranktwo{0.848} & 0.801 & 0.825 & \ranktwo{0.814} & 0.784 & 0.805 & 0.811 & 0.937 & 0.847 & 0.822 \\
\textbf{microsomes} & 0.810 & 0.785 & 0.814 & 0.901 & -- & -- & 0.804 & -- & 0.812 & 0.785 & \rankone{0.827} & -- & -- & 0.826 & 0.717 \\
\textbf{filipsPL} & 0.798 & 0.765 & 0.817 & 0.893 & 0.736 & 0.743 & 0.758 & 0.776 & -- & 0.771 & -- & 0.766 & 0.928 & 0.815 & -- \\
\textbf{Charite} & 0.785 & 0.750 & 0.811 & 0.896 & 0.688 & 0.789 & 0.739 & 0.781 & 0.751 & 0.707 & 0.798 & 0.852 & 0.880 & 0.834 & 0.700 \\
\textbf{RCC} & 0.772 & 0.751 & 0.781 & 0.872 & 0.763 & 0.747 & 0.761 & 0.792 & 0.673 & 0.781 & 0.762 & 0.755 & 0.920 & 0.795 & 0.637 \\
\textbf{frozenarm} & 0.771 & 0.759 & 0.768 & 0.865 & 0.744 & 0.722 & 0.700 & 0.740 & 0.726 & 0.745 & 0.790 & 0.752 & 0.859 & 0.803 & 0.803 \\
\textbf{ToxFit} & 0.763 & 0.753 & 0.756 & 0.862 & 0.744 & 0.757 & 0.697 & 0.738 & 0.729 & 0.729 & 0.752 & 0.689 & 0.862 & 0.803 & 0.791 \\
\textbf{CGL} & 0.759 & 0.720 & 0.791 & 0.866 & 0.742 & 0.566 & 0.747 & 0.749 & 0.737 & 0.759 & 0.727 & 0.775 & 0.880 & 0.817 & 0.738 \\
\textbf{SuperTox} & 0.743 & 0.682 & 0.768 & 0.854 & -- & 0.560 & 0.711 & 0.742 & -- & -- & -- & -- & 0.862 & 0.732 & -- \\
\textbf{kibutz} & 0.741 & 0.731 & 0.731 & 0.865 & 0.750 & 0.694 & 0.708 & 0.729 & 0.737 & 0.757 & 0.779 & 0.587 & 0.838 & 0.787 & 0.666 \\
\textbf{MML} & 0.734 & 0.700 & 0.753 & 0.871 & 0.693 & 0.660 & 0.701 & 0.709 & 0.749 & 0.750 & 0.710 & 0.647 & 0.854 & 0.815 & 0.645 \\
\textbf{NCI} & 0.717 & 0.651 & 0.791 & 0.812 & 0.628 & 0.592 & 0.783 & 0.698 & 0.714 & 0.483 & 0.703 & \rankone{0.858} & 0.851 & 0.747 & 0.736 \\
\textbf{VIF} & 0.708 & 0.702 & 0.692 & 0.827 & 0.797 & 0.610 & 0.636 & 0.671 & 0.656 & 0.732 & 0.735 & 0.723 & 0.796 & 0.648 & 0.666 \\
\textbf{Toxic Avg} & 0.644 & 0.659 & 0.607 & 0.715 & 0.721 & 0.611 & 0.633 & 0.671 & 0.593 & 0.646 & 0.640 & 0.465 & 0.732 & 0.614 & 0.682 \\
\textbf{Swamidass} & 0.576 & 0.596 & 0.593 & 0.353 & 0.571 & 0.748 & 0.372 & 0.274 & 0.391 & 0.680 & 0.738 & 0.711 & 0.828 & 0.661 & 0.585 \\
\end{tabular}
\caption{
    Results of the leading teams in the Tox21 Data Challenge, best results in bold and gray background, second best results in light gray. \emph{AVG} is the average over
    all 12 subchallenges. \emph{NR/SR} are averages over all tasks that are
    part of the ``Nuclear Receptor'' and ``Stress Response'' panels, respectively. The left side shows the
    individual sub-challenges. Team-names have been abbreviated in order to save space. The full list of
    results is available online at \url{https://tripod.nih.gov/tox21/challenge/leaderboard.jsp}.
}
\label{fig:tox21results}
\end{table*}

Our approach which was spearheaded by the deep neural network presented in this paper
showed the most consistent performance of all participants: It
never placed lower than fifth place in any of the tasks, and outright won a
total of 8 of the 15 challenges.
In particular, it achieved the best average AUC in both the SR and NR panels,
as well as as well as the best average AUC over the whole set of sub-challenges.
It was thus declared winner of both the Nuclear Receptor and the Stress Response
pannel, as well as the overall Tox21 Grand Challenge.
The detailed results are displayed in \autoref{fig:tox21results}.

\section{Conclusion}
In this paper we applied of deep neural networks to toxicity prediction.
We showed that deep networks are able to
learn a highly effective representation of chemical compounds.
In this representation we could detect toxicophores,
proven concepts which have previously often been handcrafted over decades
by experts in the field. 
It stands to reason that these representations also include novel, previously
undiscovered toxicophores that are lying dormant in the data.
Using these representations, our approach outperformed
methods that were specifically tailored for toxicological applications.

As demonstrated by the Tox21 Data Challenge, our method sets a new state of
the art in this field. As the NIH confirmed\,\cite{bib:ncats2015}, the high
quality of the models makes them suitable for deployment in leading edge
toxicological research.
We believe that Deep Learning has the ability to greatly influence
the field of toxicity prediction in the future.
Toxicology is a crucial part of modern environmental health, drug development
and pharmaceutical research, and  machine learning is on the verge of
becoming a vital part of it.


\ifdefined\isaccepted
\section*{Acknowledgments}
This work was supported in part by European Union's IAPP grant number 324554.
The authors also gratefully acknowledge the support of NVIDIA Corporation with the donation
of a GPU used for this research.
\fi


\bibliography{bibliography}

\begin{thebibliography}{35}
\providecommand{\natexlab}[1]{#1}
\providecommand{\url}[1]{\texttt{#1}}
\expandafter\ifx\csname urlstyle\endcsname\relax
  \providecommand{\doi}[1]{doi: #1}\else
  \providecommand{\doi}{doi: \begingroup \urlstyle{rm}\Url}\fi

\bibitem[Arrowsmith(2011)]{bib:Arrowsmith2011}
Arrowsmith, J.
\newblock Trial watch: phase {III} and submission failures: 2007--2010.
\newblock \emph{Nature Reviews Drug Discovery}, 10\penalty0 (2):\penalty0
  87--87, 2011.

\bibitem[Bengio(2013)]{bib:Bengio2013b}
Bengio, Y.
\newblock Deep learning of representations: Looking forward.
\newblock In \emph{Proceedings of the First International Conference on
  Statistical Language and Speech Processing}, SLSP'13, pp.\  1--37, Berlin,
  Heidelberg, 2013. Springer-Verlag.
\newblock ISBN 978-3-642-39592-5.

\bibitem[Bengio et~al.(2013)Bengio, Courville, and Vincent]{bib:Bengio2013a}
Bengio, Y., Courville, A., and Vincent, P.
\newblock Representation learning: A review and new perspectives.
\newblock \emph{IEEE Trans Pattern Anal Mach Intell}, Feb 2013.

\bibitem[Byvatov et~al.(2003)Byvatov, Fechner, Sadowski, and
  Schneider]{bib:Byvatov2003}
Byvatov, E., Fechner, U., Sadowski, J., and Schneider, G.
\newblock {Comparison of Support Vector Machine and Artificial Neural Network
  Systems for Drug/Nondrug Classification}.
\newblock \emph{Journal of Chemical Information and Computer Sciences},
  43\penalty0 (6):\penalty0 1882--1889, September 2003.

\bibitem[Cao et~al.(2013)Cao, Xu, Hu, and Liang]{bib:Cao2013b}
Cao, DS., Xu, QS., Hu, QN., and Liang, YZ.
\newblock {ChemoPy}: freely available python package for computational biology
  and chemoinformatics.
\newblock \emph{Bioinformatics}, 29\penalty0 (8):\penalty0 1092--1094, 2013.

\bibitem[Caruana(1997)]{bib:Caruana1997}
Caruana, R.
\newblock Multitask learning.
\newblock \emph{Machine Learning}, 28\penalty0 (1):\penalty0 41–75, 1997.
\newblock ISSN 0885-6125.

\bibitem[{Committee on Toxicity Testing and Assessment of Environmental Agents,
  National Research Council}(2007)]{bib:tox212007}
{Committee on Toxicity Testing and Assessment of Environmental Agents, National
  Research Council}.
\newblock \emph{Toxicity Testing in the 21st Century: A Vision and a Strategy}.
\newblock The National Academies Press, Washington, DC, 2007.
\newblock ISBN 978-0-309-15173-3.

\bibitem[Dahl et~al.(2014)Dahl, Jaitly, and Salakhutdinov]{bib:Dahl2014}
Dahl, G., Jaitly, N., and Salakhutdinov, R.
\newblock Multi-task neural networks for {QSAR} predictions.
\newblock \emph{CoRR}, abs/1406.1231, 2014.

\bibitem[Deng et~al.(2013)Deng, Li, Huang, Yao, Yu, Seide, Seltzer, Zweig, He,
  Williams, Gong, and Acero]{bib:Deng2013}
Deng, L., Li, J., Huang, JT., Yao, K., Yu, D., Seide, F., Seltzer, M., Zweig,
  G., He, Xiaodong, Williams, J., Gong, Y., and Acero, A.
\newblock Recent advances in deep learning for speech research at microsoft.
\newblock In \emph{Acoustics, Speech and Signal Processing (ICASSP), 2013 IEEE
  International Conference on}, pp.\  8604--8608, 2013.

\bibitem[Glorot et~al.(2011)Glorot, Bordes, and Bengio]{bib:Glorot2011}
Glorot, X., Bordes, A., and Bengio, Y.
\newblock Deep sparse rectifier neural networks.
\newblock In \emph{AISTATS}, pp.\  315--323, 2011.

\bibitem[Harper et~al.(2001)Harper, Bradshaw, Gittins, Green, and
  Leach]{bib:Harper2001}
Harper, G., Bradshaw, J., Gittins, J., Green, D., and Leach, A.
\newblock Prediction of biological activity for high-throughput screening using
  binary kernel discrimination.
\newblock \emph{Journal of Chemical Information and Computer Sciences},
  41\penalty0 (5):\penalty0 1295--1300, 2001.

\bibitem[Hinton et~al.(2012)Hinton, Srivastava, Krizhevsky, Sutskever, and
  Salakhutdinov]{bib:Hinton2012}
Hinton, G., Srivastava, N., Krizhevsky, A., Sutskever, I., and Salakhutdinov,
  R.
\newblock Improving neural networks by preventing co-adaptation of feature
  detectors.
\newblock July 2012.

\bibitem[Hong et~al.(2008)Hong, Xie, Ge, Qian, Fang, Shi, Su, Perkins, and
  Tong]{bib:Hong2008}
Hong, H., Xie, Q., Ge, W., Qian, F., Fang, H., Shi, L., Su, Z., Perkins, R.,
  and Tong, W.
\newblock Mold2, molecular descriptors from 2d structures for chemoinformatics
  and toxicoinformatics.
\newblock \emph{Journal of Chemical Information and Modeling}, 48\penalty0
  (7):\penalty0 1337--1344, 2008.

\bibitem[Inglese et~al.(2006)Inglese, Auld, Jadhav, Johnson, Simeonov, Yasgar,
  Zheng, and Austin]{bib:Inglese2006}
Inglese, J., Auld, D.~S., Jadhav, A., Johnson, R.~L., Simeonov, A., Yasgar, A.,
  Zheng, W., and Austin, C.~P.
\newblock {Quantitative high-throughput screening: a titration-based approach
  that efficiently identifies biological activities in large chemical
  libraries.}
\newblock \emph{Proc Natl Acad Sci U S A}, 103\penalty0 (31):\penalty0
  11473--11478, August 2006.

\bibitem[Jenkins et~al.(2007)Jenkins, Bender, and Davies]{bib:Jenkins2007}
Jenkins, J., Bender, A., and Davies, J.
\newblock In silico target fishing: Predicting biological targets from chemical
  structure.
\newblock \emph{Drug Discovery Today: Technologies}, 3\penalty0 (4):\penalty0
  413--421, 2007.

\bibitem[Kazius et~al.(2005)Kazius, McGuire, and Bursi]{bib:Kazius2005}
Kazius, J., McGuire, R., and Bursi, R.
\newblock Derivation and validation of toxicophores for mutagenicity
  prediction.
\newblock \emph{Journal of Medicinal Chemistry}, 48\penalty0 (1):\penalty0
  312--320, 2005.

\bibitem[Kier(1971)]{bib:Kier1971}
Kier, L.B.
\newblock \emph{Molecular orbital theory in drug research}.
\newblock Medicinal chemistry. Academic Press, 1971.

\bibitem[Kitchen et~al.(2004)Kitchen, Decornez, Furr, and
  Bajorath]{bib:Kitchen2004}
Kitchen, D., Decornez, H., Furr, J., and Bajorath, J.
\newblock Docking and scoring in virtual screening for drug discovery: methods
  and applications.
\newblock \emph{Nature Reviews Drug discovery}, 3\penalty0 (11):\penalty0
  935--949, 2004.

\bibitem[Kola \& Landis(2004)Kola and Landis]{bib:Kola2004}
Kola, I. and Landis, J.
\newblock Can the pharmaceutical industry reduce attrition rates?
\newblock \emph{Nat Rev Drug Discov}, 3\penalty0 (8):\penalty0 711--716, August
  2004.
\newblock ISSN 1474-1776.

\bibitem[Krizhevsky et~al.(2012)Krizhevsky, Sutskever, and
  Hinton]{bib:Krizhevsky2012}
Krizhevsky, A., Sutskever, I., and Hinton, G.
\newblock Imagenet classification with deep convolutional neural networks.
\newblock In Pereira, F., Burges, C.J.C., Bottou, L., and Weinberger, K.Q.
  (eds.), \emph{Advances in Neural Information Processing Systems 25}, pp.\
  1097--1105. Curran Associates, Inc., 2012.

\bibitem[Lin(2000)]{bib:Lin2000}
Lin, SK.
\newblock Pharmacophore perception, development and use in drug design. edited
  by osman f. g{\"u}ner.
\newblock \emph{Molecules}, 5\penalty0 (7):\penalty0 987--989, 2000.
\newblock ISSN 1420-3049.

\bibitem[Lowe et~al.(2011)Lowe, Mussa, Mitchell, and Glen]{bib:Lowe2011}
Lowe, R, Mussa, H., Mitchell, J., and Glen, R.
\newblock Classifying molecules using a sparse probabilistic kernel binary
  classifier.
\newblock \emph{Journal of Chemical Information and Modeling}, 51\penalty0
  (7):\penalty0 1539--1544, 2011.

\bibitem[Mussa et~al.(2013)Mussa, Mitchell, and Glen]{bib:Mussa2013}
Mussa, H., Mitchell, J., and Glen, R.
\newblock {Full "Laplacianised" posterior naive Bayesian algorithm}.
\newblock \emph{Journal of Cheminformatics}, 5\penalty0 (1):\penalty0 37+,
  August 2013.

\bibitem[Nair \& Hinton(2010)Nair and Hinton]{bib:Nair2010}
Nair, V. and Hinton, G.
\newblock Rectified linear units improve restricted boltzmann machines.
\newblock In \emph{Proceedings of the 27th International Conference on Machine
  Learning (ICML)}, pp.\  807--814, 2010.

\bibitem[{National Center for Advancing Translational
  Sciences}(2014)]{bib:tox21challenge}
{National Center for Advancing Translational Sciences}.
\newblock {Tox21 Data Challenge 2014}.
\newblock \url{https://tripod.nih.gov/tox21/challenge/}, 2014.
\newblock [Online; last accessed 05-January-2015].

\bibitem[{National Center for Advancing Translational
  Sciences}(2015)]{bib:ncats2015}
{National Center for Advancing Translational Sciences}.
\newblock {NCATS Announces Tox21 Data Challenge Winners}.
\newblock
  \url{http://www.ncats.nih.gov/news-and-events/features/tox21-challenge-winners.html},
  2015.
\newblock [Online; last accessed 06-February-2015].

\bibitem[Nigsch et~al.(2008)Nigsch, Bender, Jenkins, and
  Mitchell]{bib:Nigsch2008}
Nigsch, F., Bender, A., Jenkins, J., and Mitchell, J.
\newblock Ligand-target prediction using winnow and naive bayesian algorithms
  and the implications of overall performance statistics.
\newblock \emph{Journal of Chemical Information and Modeling}, 48\penalty0
  (12):\penalty0 2313--2325, 2008.

\bibitem[R. et~al.(2012)R., Y., F., Glen, and Mitchell]{bib:Lowe2012}
R., Lowe, Y., Mussa, F., Nigsch, Glen, R., and Mitchell, J.
\newblock Predicting the mechanism of phospholipidosis.
\newblock \emph{Journal of Cheminformatics}, 4\penalty0 (1):\penalty0 2, 2012.

\bibitem[{Ramsundar} et~al.(2015){Ramsundar}, {Kearnes}, {Riley}, {Webster},
  {Konerding}, and {Pande}]{bib:Ramsundar2015}
{Ramsundar}, B., {Kearnes}, S., {Riley}, P., {Webster}, D., {Konerding}, D.,
  and {Pande}, V.
\newblock {Massively Multitask Networks for Drug Discovery}.
\newblock \emph{CoRR}, abs/1502.02072, 2015.

\bibitem[Rogers \& Hahn(2010)Rogers and Hahn]{bib:Rogers2010}
Rogers, D. and Hahn, M.
\newblock Extended-connectivity fingerprints.
\newblock \emph{Journal of Chemical Information and Modeling}, 50\penalty0
  (5):\penalty0 742--754, May 2010.

\bibitem[Srivastava et~al.(2014)Srivastava, Hinton, Krizhevsky, Sutskever, and
  Salakhutdinov]{bib:Srivastava2014}
Srivastava, N., Hinton, G., Krizhevsky, A., Sutskever, I., and Salakhutdinov,
  R.
\newblock Dropout: A simple way to prevent neural networks from overfitting.
\newblock \emph{Journal of Machine Learning Research}, 15:\penalty0 1929--1958,
  2014.

\bibitem[Unterthiner et~al.(2014)Unterthiner, Mayr, Klambauer, Steijaert,
  Wegner, Ceulemans, and Hochreiter]{bib:unterthiner2014}
Unterthiner, T., Mayr, A., Klambauer, G., Steijaert, M., Wegner, J.~K.,
  Ceulemans, H., and Hochreiter, S.
\newblock Deep learning as an opportunity in virtual screening.
\newblock In \emph{Deep Learning and Representation Learning Workshop, {NIPS
  2014}}, Montreal, Canada, Dec 2014.

\bibitem[Verbist et~al.(2015)Verbist, Klambauer, Vervoort, Talloen, Shkedy,
  Thas, Bender, G{\"o}hlmann, and Hochreiter]{bib:Klambauer2015}
Verbist, B., Klambauer, G., Vervoort, L., Talloen, W., Shkedy, Z., Thas, O.,
  Bender, A., G{\"o}hlmann, H., and Hochreiter, S.
\newblock Using transcriptomics to guide lead optimization in drug discovery
  projects: Lessons learned from the {QSTAR} project.
\newblock \emph{Drug Discovery Today}, 2015.
\newblock ISSN 1359-6446.

\bibitem[Woods et~al.(2007)Woods, Vanden~H., and Rusyn]{bib:Woods2007}
Woods, C.~G., Vanden~H., John~P., and Rusyn, I.
\newblock Genomic profiling in nuclear receptor-mediated toxicity.
\newblock \emph{Toxicologic Pathology}, 35\penalty0 (4):\penalty0 474--494,
  2007.

\bibitem[Xia et~al.(2004)Xia, Maliski, Gallant, and Rogers]{bib:Xia2004}
Xia, X., Maliski, E., Gallant, P., and Rogers, D.
\newblock {Classification of Kinase Inhibitors Using a Bayesian Model}.
\newblock \emph{Journal of Medicinal Chemistry}, 47\penalty0 (18):\penalty0
  4463--4470, August 2004.

\end{thebibliography}
\bibliographystyle{icml2015}

\end{document}